\documentclass[runningheads]{llncs}

\usepackage[mobile]{eccv}

\usepackage{eccvabbrv}

\usepackage{graphicx}
\usepackage{booktabs}

\usepackage[accsupp]{axessibility}  % Improves PDF readability for those with disabilities.

\usepackage{epsfig}
\usepackage{graphicx}
\usepackage{amsmath}
\usepackage{amssymb}
\usepackage{bm}
\usepackage{float}
\usepackage{tabularx}
\usepackage{microtype}
\usepackage{wrapfig}

\usepackage{subfloat}
\usepackage{booktabs} % for professional tables
\usepackage{url}
\usepackage{caption}
\usepackage{multirow}
\usepackage{multicol}
\usepackage{array}
\usepackage{pifont}
\usepackage{xcolor}
\usepackage{booktabs}
\usepackage[normalem]{ulem}
\usepackage{xspace}
\usepackage{diagbox}
\usepackage{color}
\usepackage{enumitem}
\usepackage{bm}
\usepackage{algorithm}
\usepackage{listings}
\usepackage{xspace}

\usepackage{etoolbox}
\makeatletter
\AfterEndEnvironment{algorithm}{\let\@algcomment\relax}
\AtEndEnvironment{algorithm}{\kern2pt\hrule\relax\vskip3pt\@algcomment}
\let\@algcomment\relax
\newcommand\algcomment[1]{\def\@algcomment{\footnotesize#1}}
\renewcommand\fs@ruled{\def\@fs@cfont{\bfseries}\let\@fs@capt\floatc@ruled
  \def\@fs@pre{\hrule height.8pt depth0pt \kern2pt}%
  \def\@fs@post{}%
  \def\@fs@mid{\kern2pt\hrule\kern2pt}%
  \let\@fs@iftopcapt\iftrue}
\makeatother
\newcommand{\stab}{Stabilizer\xspace}
\newcommand{\boo}{Booster\xspace}

\newcommand\blfootnote[1]{\begingroup\renewcommand\thefootnote{}\footnote{#1}\addtocounter{footnote}{-1}\endgroup}

\DeclareMathOperator*{\argmax}{arg\,max}

\usepackage[pagebackref,breaklinks,colorlinks,citecolor=eccvblue]{hyperref}
\usepackage{orcidlink}

\author{%
  Haoran Chen$^{1}$
  \and
  Zuxuan Wu$^{1,2\dagger}$ 
    \and
  Xintong Han$^{3}$ 
    \and \\
  Menglin Jia$^{4}$
    \and
  Yu-Gang Jiang$^{1,2}$
\\ 
}

\authorrunning{H. Chen et al.}
\institute{$^{1}$Shanghai Key Lab of Intell. Info. Processing, School of CS, Fudan University \\
$^{2}$Shanghai Collaborative Innovation Center of Intelligent Visual Computing \\
$^{3}$Huya Inc~\hspace{10pt} $^{4}$Cornell Univeristy}

\begin{document}

\title{PromptFusion: Decoupling Stability and Plasticity for Continual Learning}

\maketitle

\begin{abstract}
  Current research on continual learning mainly focuses on relieving catastrophic forgetting, and most of their success is at the cost of limiting the performance of newly incoming tasks. Such a trade-off is referred to as the stability-plasticity dilemma and is a more general and challenging problem for continual learning. However, the inherent conflict between these two concepts makes it seemingly impossible to devise a satisfactory solution to both of them simultaneously. Therefore, we ask, ``is it possible to divide them into two separate problems to conquer them independently?''. To this end, we propose a prompt-tuning-based method termed PromptFusion to enable the decoupling of stability and plasticity. Specifically, PromptFusion consists of a carefully designed \stab module that deals with catastrophic forgetting and a \boo module to learn new knowledge concurrently. Furthermore, to address the computational overhead brought by the additional architecture, we propose PromptFusion-Lite which improves PromptFusion by dynamically determining whether to activate both modules for each input image. Extensive experiments show that both PromptFusion and PromptFusion-Lite achieve promising results on popular continual learning datasets for class-incremental and domain-incremental settings. Especially on Split-Imagenet-R, one of the most challenging datasets for class-incremental learning, our method can exceed state-of-the-art prompt-based methods by more than 5\% in accuracy, with PromptFusion-Lite using 14.8\% less computational resources than PromptFusion. Code is available at \href{https://github.com/HaoranChen/PromptFusion}{https://github.com/HaoranChen/PromptFusion}. 
  \blfootnote{$^{\dagger}$ Corresponding author.}
  \keywords{Continual Learning \and Prompt Tuning}

\end{abstract}

\section{Introduction}
Despite great advances in deep learning, neural networks are often trained in a static supervised manner where all training data are available at once \cite{alexnet, Resnet, fasterrcnn}. Continual learning \cite{continualsurvey1, continualsurvey2, continualsurvey3}, on the contrary, studies the behavior of neural networks under a more realistic scenario in which data arrive in a continual and 
\setlength{\intextsep}{5pt}
\begin{wrapfigure}[14]{r}{0.52\textwidth}
   \includegraphics[width=0.52\columnwidth,height=0.25\columnwidth]{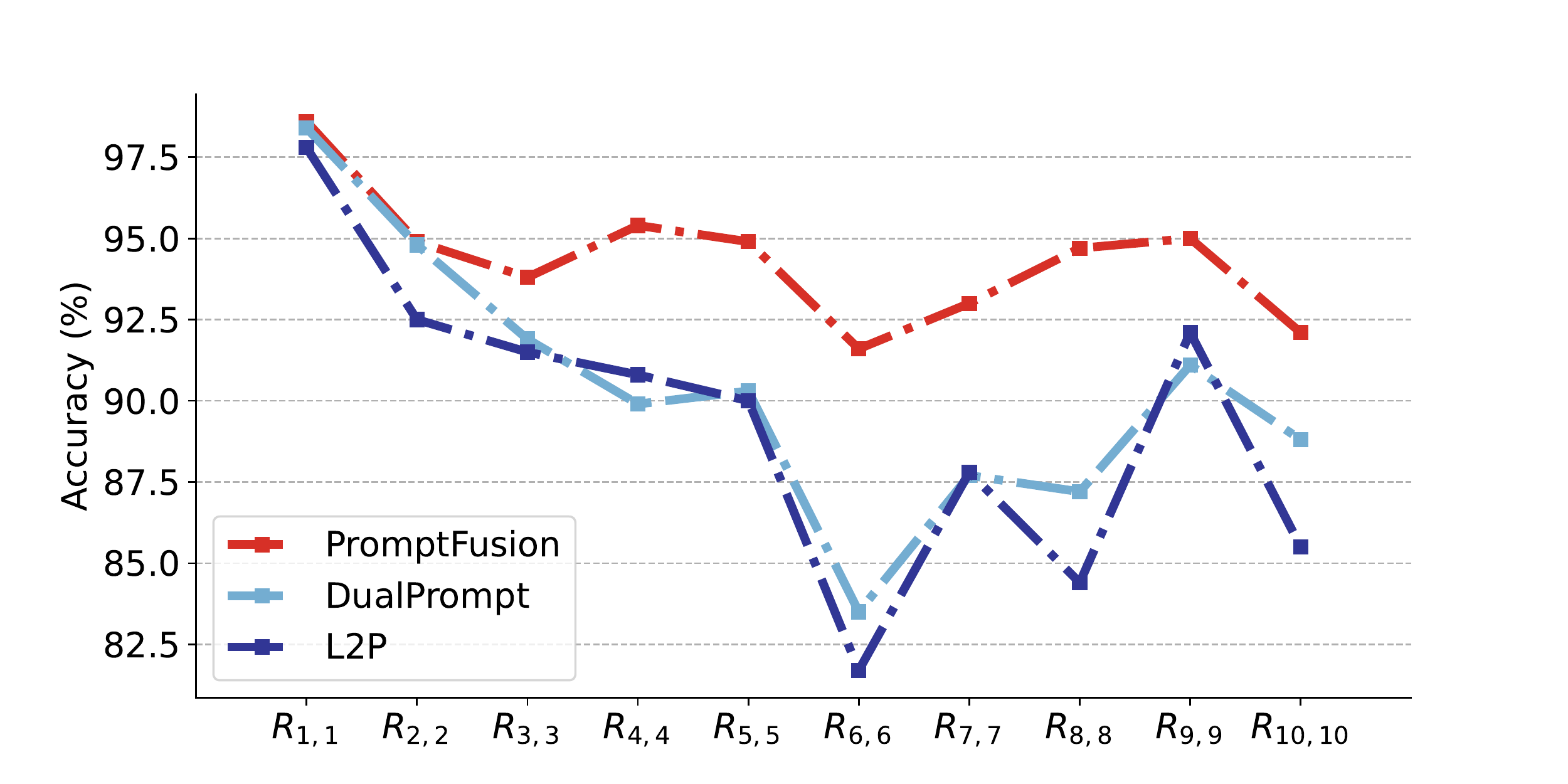}
   \caption{Performance on the most recently learned task on the Split-Cifar100 dataset. Here, $R_{T,i}$ is the test accuracy on task $i$ after learning task $T$.
   As is shown, the plasticity of L2P~\cite{learningtoprompt} and DualPrompt~\cite{wang2022dualprompt} is limited. 
   }
   \label{fig:l2p_dualprompt_plasticity}

\end{wrapfigure}
dynamic procedure. Ideally, when confronted with new data, state-of-the-art models should be both stable to prevent performance degradation for previous tasks and plastic to learn sufficient information for the new task \cite{wu2021striking}. However, in practice, it is hard or even impossible to maintain such a balance, which is known as the stability-plasticity dilemma \cite{abraham2005memory, mermillod2013stability}. As a result, most of the current literature mainly focuses on one side of the problem, \ie, relieving the problem of catastrophic forgetting, but overlooking the unseen sacrifice on the other side. A typical example is regularization-based approaches such as EWC~\cite{ewc}. In EWC, important parameters for previous tasks are kept intact which inevitably limits the capacity of learning new knowledge.

 Since directly balancing this trade-off is extremely challenging, in this work, we instead seek to address the issue from a different perspective. Specifically, inspired by the Complementary Learning System~\cite{cls, kumaran2016learning}, a biology theory that suggests intelligent agents must possess two learning systems, we hypothesize that the dilemma can be resolved similarly by leveraging two different architectures to tackle stability and plasticity independently.  However, the incorporation of an additional architecture raises computational concerns as it might severely complicate the optimization process. So now the question becomes, is there a computationally efficient way to implement this idea?

 Recently, prompt tuning \cite{liu2021promptsurvey, lester2021power, wei2021pretrained} has become an emerging trend of fine-tuning models on downstream tasks in a computationally efficient manner. Since prompt tuning only trains an additional small set of parameters and freezes the pretrained backbone, it gives a potential solution to solve the above question. In fact, due to its strong transferring ability, prompt tuning has already been explored in continual learning~\cite{learningtoprompt, wang2022dualprompt, wang2022sprompt}. The first work to do so trains a prompt pool followed by optimizing a prompt selection process for each task and achieves remarkable performances. Consequently, various follow-up approaches are introduced. However, as depicted in Fig~\ref{fig:l2p_dualprompt_plasticity}, their plasticity is severely limited.

To this end, we propose PromptFusion, a simple yet effective framework that decouples stability and plasticity for continual learning using a \stab module and a \boo module. Specifically, we instantiate the \stab with CoOp~\cite{coop} and the \boo with VPT~\cite{vpt}, two mainstream prompt tuning methods. For the \stab, PromptFusion initializes a new set of prompts for each incoming task and concatenates it to previously learned ones. During training, prompts from past tasks are frozen and only the current task-specific set is trained. As for the \boo, PromptFusion trains the same set of prompts for all tasks. Therefore, since only the newly added prompts of the \stab are involved in training, historical information can be well preserved (\emph{stability}). In contrast, all prompts of the \boo are continuously updated, thereby allowing full learning of new information (\emph{plasticity}). As a result, our design makes the \stab suitable for stability and the \boo suitable for plasticity, achieving the best of both worlds.

To justify such an approach, a pilot study is carried out to empirically analyze the continual learning ability of both modules. Surprisingly, we find that in addition to aligning with our intention, their performance diverges on different datasets. Specifically, the \stab is discovered to be much more robust to intra-class variations and is thus more suitable for the domain-incremental setting and complex datasets such as ImageNet-R, while the \boo is proficient on simple ones such as Cifar100. In light of such variation, to make full use of both modules, we train a weight parameter $\lambda$ conditioned on the training dataset for the ensemble of their output logits. Following most continual learning approaches~\cite{icarl, il2m}, we also apply a learnable weight mask to the final output logits for accommodating the imbalanced class distribution under both rehearsal and rehearsal-free settings (details in the following section). Extensive experiments on both class-incremental and domain-incremental datasets show that our proposed method achieves state-of-the-art results. 
 
 Nevertheless, while PromptFusion results in impressive performances, naively using both modules simultaneously would inevitably increase computational costs. Since existing works have already demonstrated the capability of using either the \stab or the \boo in continual learning tasks~\cite{wang2022dualprompt, wang2022sprompt, learningtoprompt}, we wonder if we can use one of them as a base module that runs by default, and have the model adaptively decides on-the-fly whether to use the other one on a per-input basis. To this end, we further propose a computationally efficient framework PromptFusion-Lite that empirically chooses the \stab as the default module and applies the Gumbel-Softmax trick~\cite{gumbelsoftmax} to determine whether or not to activate the \boo conditioned on inputs. Surprisingly, PromptFusion-Lite can still achieve competitive performance on continual learning benchmarks, while effectively dropping the overall computational overhead. In summary, our contributions are three-fold:
 \begin{itemize}
\item[$\bullet$] We introduce PromptFusion and PromptFusion-Lite to address the stability-plasticity dilemma of continual learning in a computationally efficient manner. They take advantage of two modules, the \stab and the \boo, which decouple stability and plasticity into two independent problems. PromptFusion naively uses both modules simultaneously, while PromptFusion-Lite adaptively selects the appropriate modules for each input. 
\item[$\bullet$]We conduct a detailed analysis of the \stab and the \boo modules in the continual learning setting and discover that their performance diverges on different datasets. In particular, the \stab is robust to intra-class variations and thus performs better with complex datasets in domain-incremental learning settings, while the \boo performs well only in class-incremental learning settings.
\item[$\bullet$] Both our proposed PromptFusion and PromptFusion-Lite achieve state-of-the-art results on several popular benchmarks for class-incremental and domain-incremental learning. Specifically, on Imagenet-R, the most challenging class-incremental learning benchmark, PromptFusion achieves an average accuracy of 78.7\% in the memory-free condition, surpassing the state-of-the-art method CODA-Prompt by 5.3\%. PromptFusion-Lite can achieve a competing accuracy of 75.6\% with 14.8\% less computational overhead.
\end{itemize}

\section{Related Work}
\subsection{Continual Learning}
In continual learning, the principle is to train a model on several tasks sequentially without forgetting knowledge from previously learned ones. There are three fundamental settings in literature for continual learning, namely task incremental~\cite{continualsurvey1, ewc, icarl}, class incremental~\cite{douillard2022dytox, yan2021dynamically}, and domain incremental~\cite{fini2022self, van2019three}. For task incremental learning, the task identity for the input sample is provided at test time and therefore is regarded as the most relaxed condition. On the contrary, class incremental and domain incremental learning treat all samples the same during inference without prior knowledge about the task identity. The difference between these two is that in class incremental learning, data for each task are generally from the same distribution but belong to different categories. However, in domain incremental learning, data for each task belong to different distributions but have the same class labels. 

Throughout the years, numerous efforts have been devoted to tackling continual learning, which can be mainly categorized into three groups: rehearsal-based methods~\cite{il2m, icarl, lopez2017gradient, shin2017generativereplay, hou2019lucir} where memory is utilized to store samples from the past, architecture-based methods~\cite{douillard2022dytox, yan2021dynamically, cpg, progressive, von2019hyper, mirzadeh2022architecture, mallya2018packnet} where the network expands for an incoming task, and regularization based methods~\cite{ewc, lwf, mas, si, lwm} where important parameters for previous tasks remain unchanged. However, most of these approaches mainly focus on the problem of catastrophic forgetting without considering the performance of learning the new task.

\subsection{Prompt Learning}
Recently, researchers in NLP have shown that learned large-scale language models can handle a wide range of downstream tasks with only a few or even no samples by prepending instructions to the input text~\cite{li2021prefix, liu2021promptsurvey, lester2021power}. Such instruction texts are called prompts. Consequently, prompts can be tuned instead of the weights of the entire network for a more efficient adaptation to downstream tasks. The success of prompt learning in NLP has also garnered attention in the vision community that motivates the establishment of many related methods~\cite{ju2021videoprompting, wu2024building, vpt, coop, lu2022promptdistribution, bayesianpromptlearning}. For example, DAPL~\cite{dapl} applies prompt learning in unsupervised domain adaptation by training a prompt for each source-target domain pair. MPA~\cite{mpa} extends it by adapting to the multi-source scenario through a two-stage alignment process. In the context of continual learning, a series of prompt-based works has achieved tremendous success. In L2P~\cite{learningtoprompt} and DualPrompt~\cite{wang2022dualprompt}, a prompt pool is trained such that for each task, the model samples task-specific prompts from it using a key-value selection process. However, they cannot be trained end-to-end, as the keys are optimized locally. Furthermore, they assume that every data in the mini-batch uses the same set of prompts, which is problematic during inference as data from different tasks might be present for the same mini-batch. Similarly, S-Prompt~\cite{wang2022sprompt} is proposed in a similar fashion built on top of CoOp. It is, however, specifically designed for domain-incremental learning.

\section{Preliminary}
\label{sec:preliminary}
Since we instantiate PromptFusion with CoOp and VPT, two types of prompt learning approaches, we first briefly review the two methods, followed by presenting in detail how they are leveraged in the \stab and the \boo modules. 

\noindent\textbf{CoOp}
CoOp~\cite{coop} is a large-scale vision-language representation learning model built on top of CLIP~\cite{radford2021clip}. It consists of an image encoder $f$ and a text encoder $g$ that aligns input images with text prompts. Unlike CLIP where the prompts are usually in the form of ``a photo of a [CLS]", prompts in CoOp are trainable token parameters $V_i$ of length $M$ in the form of ``$[V_1]...[V_M]$[CLS]". Given an image $\bm{x}$ with its label $y$ and text prompt $\bm{P}_k$ for class $k$, CoOp first maps them to the same embedding space. Then they are aligned in a contrastive manner such that:
\begin{equation} 
\label{eqn:clip}
p(y=k|\bm{x}) = \frac{\text{exp}(<g(\bm{P}_k),f(\bm{x})>/T)}{\sum_{i=1}^{K}\text{exp}(<g(\bm{P}_i),f(\bm{x})>/T)}
\end{equation}
is maximized when the input image $\bm{x}$  belongs to class $k$. Here, $K$ is the total number of classes, $<\cdot, \cdot>$ denotes the cosine similarity, and $T$ is a temperature parameter. During inference, the predicted label of a test image $\bm{x}$ is:
\begin{equation}
    \argmax_{k} <g(\bm{P}_k),f(\bm{x})>, k \in \{1, ..., K\}.
\end{equation}

\noindent\textbf{VPT}
In contrast to CoOp, where the model seeks alignment from two modalities, VPT only relies on a Vision Transformer (ViT) that leverages prompt learning in a pure vision fashion~\cite{vpt}. In VPT, an input image is first divided into $m$ fixed-sized patches $I_i$. With a class token, the input is first embedded into a latent space with positional embedding. Then, learnable prompt tokens $U_i$ of length $p$ are attached to the input in the form ``[CLS]$ [U_1]...[U_p][I_1]...[I_m]$". In shallow VPT, prompts are only inserted to inputs for the first Transformer layer, while for deep VPT, prompts are inserted for every layer. In this work, shallow VPT is adopted.

\noindent\textbf{Prompt Tuning for Continual Learning} In the present study, we focus on both the class incremental and the domain incremental setting. Formally, given $N$ tasks $\mathcal{T} = (T_1, T_2, ..., T_N)$ where data for task $T_t$ is denoted as $\mathcal{D}(\mathcal{X}^{t}, \mathcal{Y}^{t})$, in class incremental scenarios, $\mathcal{Y}^{i} \cap  \mathcal{Y}^{j} = \emptyset$ with $P(\mathcal{X}^i) = P(\mathcal{X}^{j})$, while for domain incremental scenarios, $\mathcal{Y}^{i} = \mathcal{Y}^{j}$ with $P(\mathcal{X}^i) \neq P(\mathcal{X}^{j})$, $i, j \in \{1, ..., N\}$, and $i \neq j$. Here $P(\cdot)$ denotes the probability density function. Since our method incorporates the usage of rehearsal memories, during each training phase, the model has access to both data from the current task and a few stored past samples, and the goal is to continuously update the model without forgetting past knowledge (\emph{stability}) while achieving good results for the current one (\emph{plasticity}).

One difference between VPT and CoOp is that prompts in CoOp are class-dependent while VPT has no such restrictions. Therefore, for incoming new classes, VPT is allowed to reuse the same set of prompts while CoOp, on the other hand, must learn new ones. As a result, for each task $T_t$, a new set of text prompts $\bm{P}_t^{stab} \in \mathbb{R}^{\frac{K}{N} \times M \times e^{stab}}$ is initialized for CoOp, where $\frac{K}{N}$ is the number of classes in each task and $e^{stab}$ is the embedding dimension. If $t > 1$, $\bm{P}_t^{stab}$ is concatenated with previously learned prompts $\bm{P}^{stab} = \text{Concat}[\bm{P}_1^{stab}, ..., \bm{P}_t^{stab}]$ for alignment with image features in the embedding space. Note that $\bm{P}_1^{stab},..., \bm{P}_{t-1}^{stab}$ is kept frozen. As for VPT, another set of prompts $\bm{P}^{boost} \in \mathbb{R}^{p \times e^{boost}}$ is initialized before the training of the first task $T_1$ and constantly updated for each task $T_t$. While we could have trained a prompt for each task using VPT and concatenated it with previously learned ones as well, experiments show that such an approach results in poor performance. Thus, we stick to the above-stated design.

\begin{figure*}[t]
  \centering
   \includegraphics[width=\linewidth]{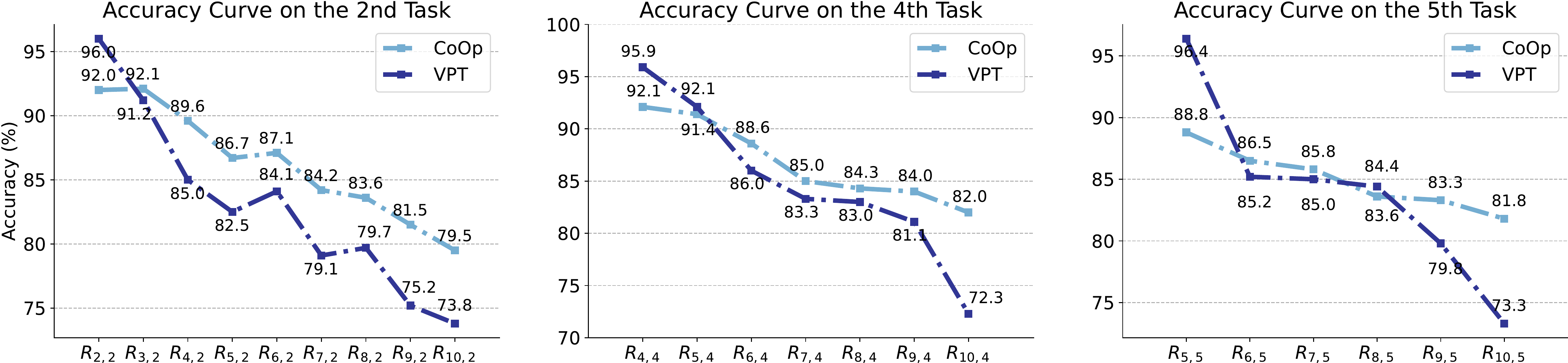}
   \caption{Stability comparison between CoOp and VPT. Accuracy curves for tasks $T_2$, $T_4$, and $T_5$ using the two modules are presented. All three graphs show a similar trend where performance degradation in CoOp is much smaller than that in VPT. This shows that CoOp is much more robust against forgetting than VPT.}
   \label{fig:stability_cv}
\end{figure*}

\section{Pilot Study}
\label{sec:pilot study}
\setlength{\intextsep}{0pt}
\begin{wrapfigure}[12]{r}{0.35\textwidth}
   \includegraphics[width=0.35\columnwidth,height=0.3\columnwidth]{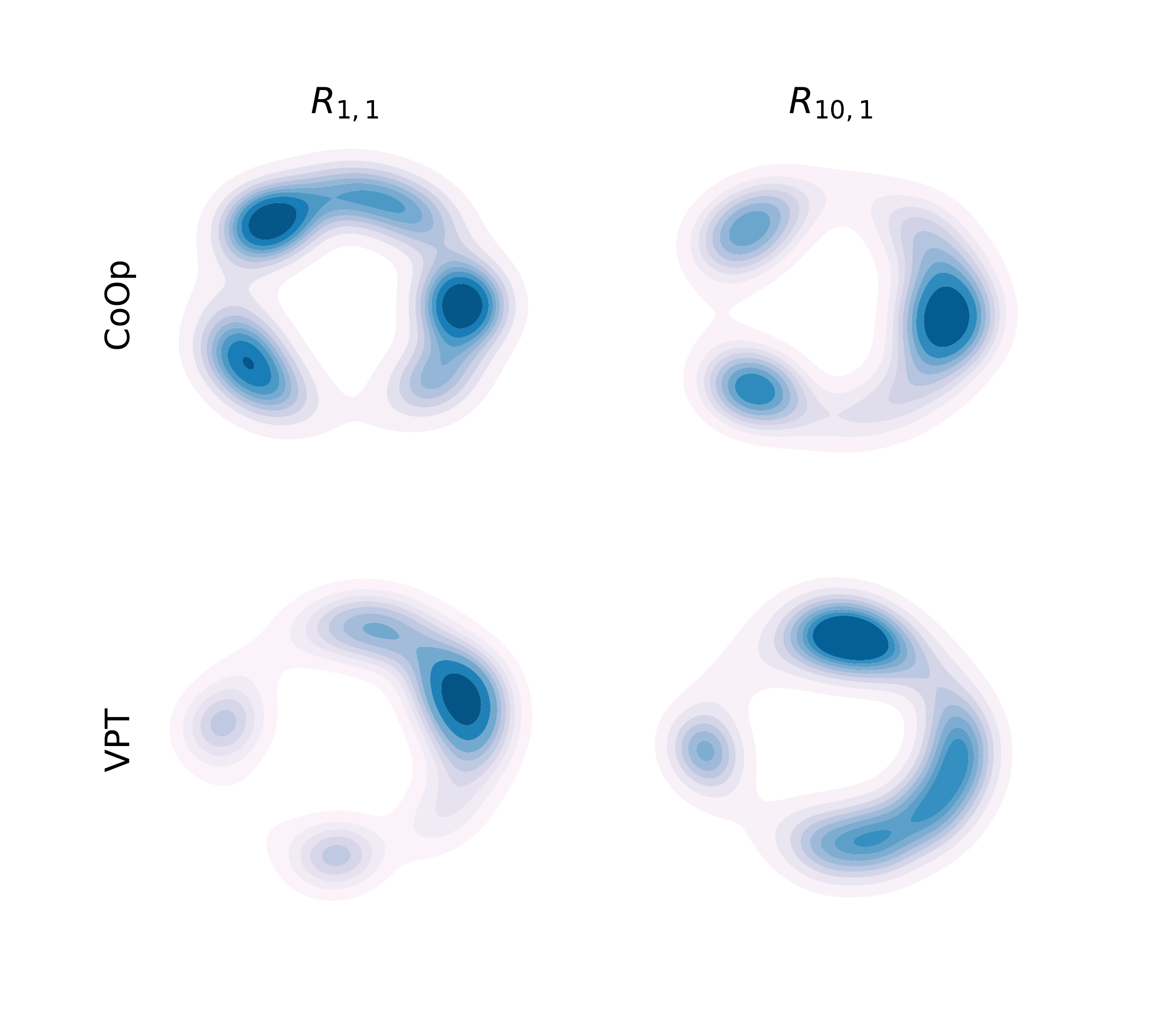}
  \caption{KDE analysis on task $T_1$.
  }
   \label{fig:kde_cv}
\end{wrapfigure}

In this section, we conduct a pilot study exploring whether stability and plasticity can be decoupled by leveraging the two proposed modules, the \stab and the \boo. To this end, we systematically analyze the performance of CoOp and VPT in the setting of continual learning.

\begin{figure}[t]
    \centering
    % First minipage for the first figure
    \begin{minipage}{.48\textwidth}
        \centering
        \includegraphics[width=\linewidth]{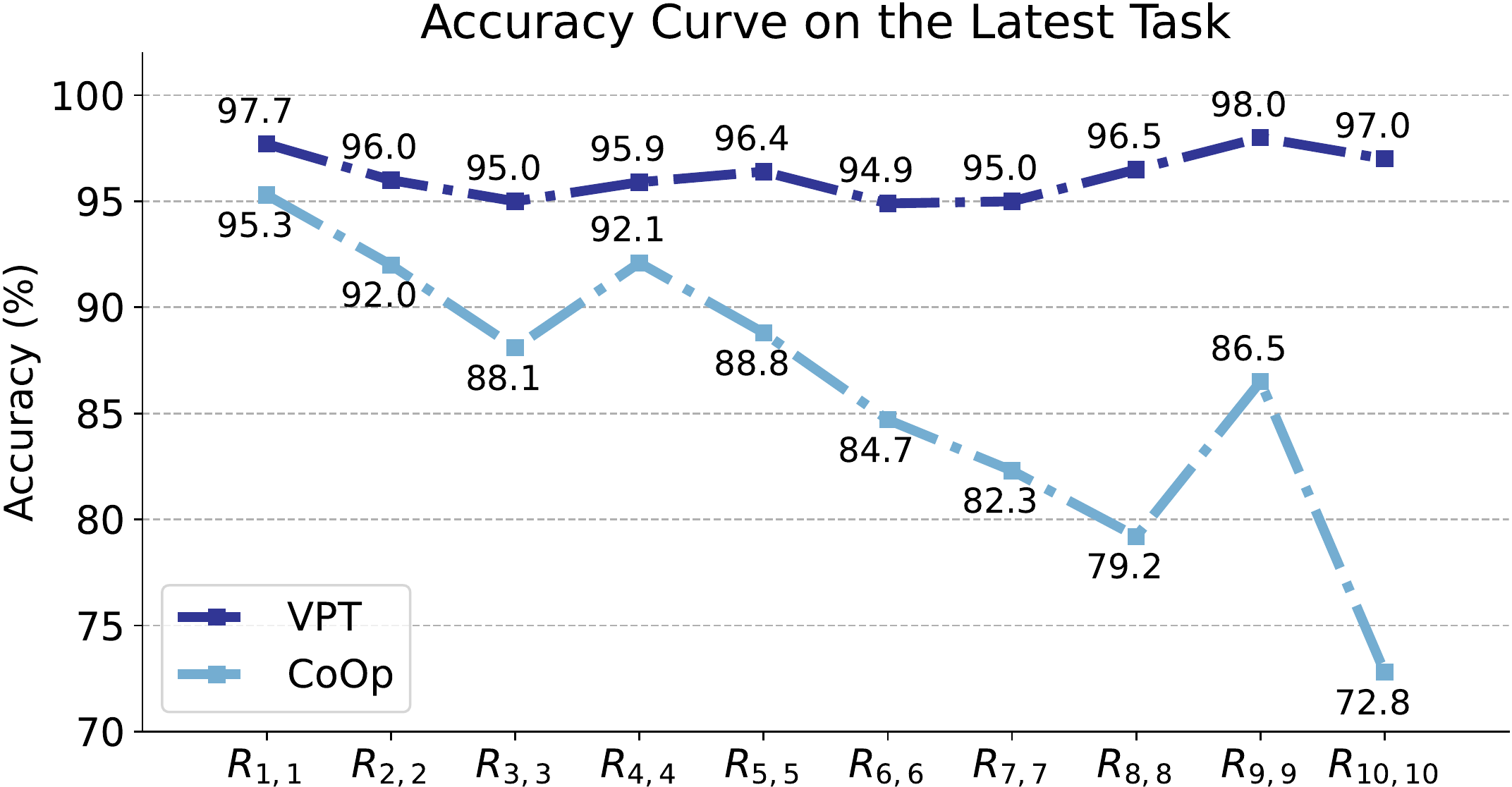}
        \caption{Plasticity comparison between CoOp and VPT, where they exhibit opposite patterns.}
        \label{fig:plasticity_cv}
    \end{minipage}\hfill % Space between the two minipages
    % Second minipage for the second figure
    \begin{minipage}{.48\textwidth}
        \centering
        \includegraphics[width=\linewidth]{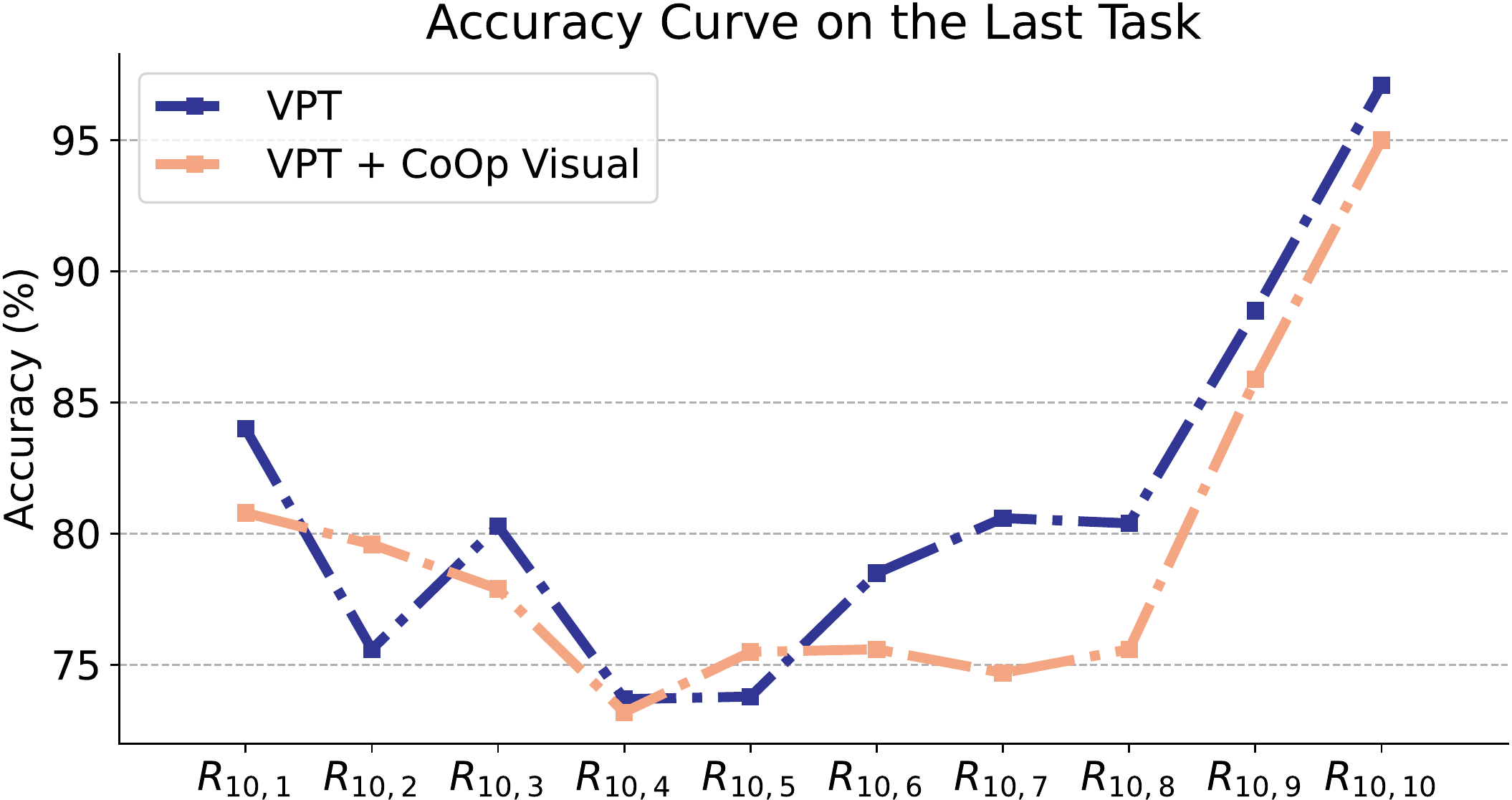}
        \caption{Comparison between VPT and VPT with weights from CoOp's visual encoder, \ie, CLIP.}
        \label{fig:coop_visual}
    \end{minipage}
\end{figure}

\noindent\textbf{Decoupling Stability and Plasticity}  We begin the analysis by justifying our statement that our design of CoOp is suitable for stability while VPT is suitable for plasticity. Figure~\ref{fig:stability_cv} presents three accuracy curves for task $T_2, T_4$ and $T_5$ on the Split-Cifar100 dataset. As shown, CoOp suffers much less from forgetting with an average performance drop of 9.9\%; whereas the average drop for VPT reaches 23.0\%, indicating that the stability for VPT is limited. We also plot the change in feature distribution for task $T_1$ in Figure~\ref{fig:kde_cv} with Gaussian
Kernel Density Estimation (KDE). KDE is a popular non-parametric method for estimating the probability density function based on kernels as weights. As is shown in the graph, the distribution change for CoOp compared to VPT is much smaller, which coincides with results in Figure~\ref{fig:stability_cv}. Figure~\ref{fig:plasticity_cv}, on the other hand, depicts the performance for each newly learned task. While accuracy for VPT consistently achieves around 95\%, CoOp exhibits a clear pattern of deterioration. Their margin in performance reaches up to 24.8\% at $T_{10}$ and could get worse when more tasks follow. While the evidence from both figures supports our hypothesis, an additional crucial piece is required to complete the entire picture: CoOp and VPT differ in their backbone networks.

\noindent\textbf{Prompt or Backbone?}
To test whether patterns in Figure~\ref{fig:stability_cv} and Figure~\ref{fig:plasticity_cv} are indeed from different designs of prompts, an additional set of experiments is conducted where we apply pre-trained weights of the image encoder of CoOp to VPT. As is shown in Figure~\ref{fig:coop_visual}, experimental results demonstrate that their performances remain approximately the same. They both demonstrate a preference for plasticity over stability, as the performance on the last task $T_{10}$ is high. Therefore, we conclude that the effect of the backbone network weight on the patterns we discovered is neglectable.

Furthermore, another interesting finding from Figure~\ref{fig:coop_visual} is that using the backbone weights from the visual encoder of CoOp, i.e., CLIP~\cite{radford2021clip},  results in a worse performance. This is surprising as CLIP is generally considered more powerful than ImageNet pre-trained ViT. As a matter of fact, \cite{wang2022dualprompt} reports a similar trend in performance drop when switching state-of-the-art continual learning methods with a stronger backbone, suggesting that a stronger network trained in a fully supervised manner does not necessarily lead to better continual learning ability.

\begin{figure}[h]
  \centering
   \includegraphics[width=0.6\columnwidth,height=0.33\columnwidth]{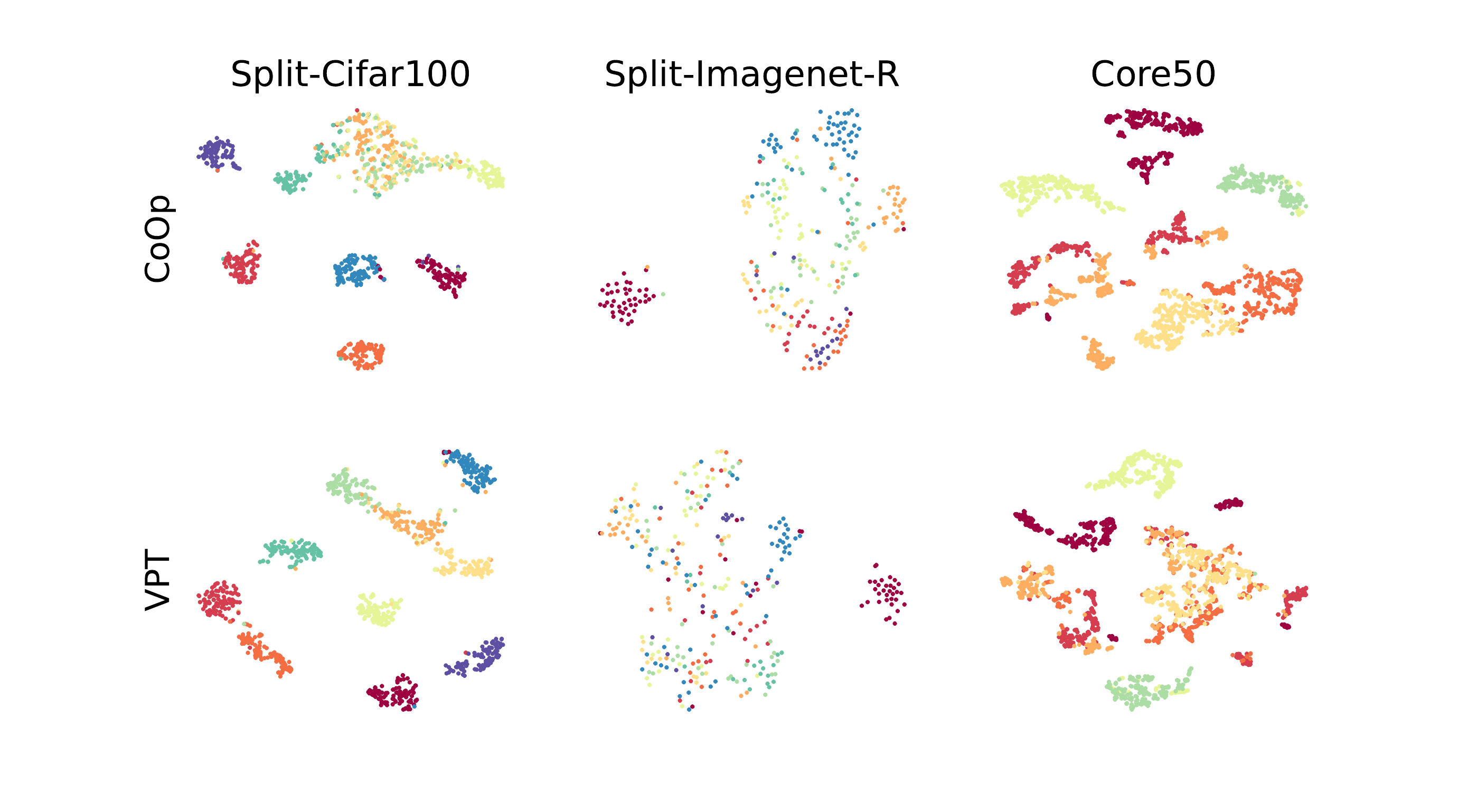}
   \caption{t-SNE visualization on Split-Cifar100, Split-Imagenet-R, and Core50. }
   \label{fig:tsne}
\end{figure}

\noindent\textbf{Divergent Performance} Meanwhile, while testing both modules on different types of datasets, we find out that CoOp is more advanced on complex datasets especially when large intra-class variations exist. VPT, on the other hand, handles simpler ones better. To justify this finding, a t-SNE visualization on Split-Cifar100, Split-Imagenet-R, and Core50 is presented in Figure~\ref{fig:tsne}. Here, Split-Cifar100 is a relatively simple dataset, while Split-Imagenet-R and Core50 incorporate covariate shifts in the data distribution. As depicted in Figure~\ref{fig:tsne}, CoOp exhibits a more clustered feature than VPT on Imagenet-R and Core-50, suggesting that it is more robust to intra-class variations. Alternatively, on Cifar100, the feature obtained from VPT is more clustered. Therefore, by utilizing the two modules, another potential benefit is that the resulting model can accommodate the varying characteristics of different datasets.

\section{Method}
\label{sec:prompt fusion}

\begin{figure}[t!]
  \centering
   \includegraphics[width=0.93\linewidth]{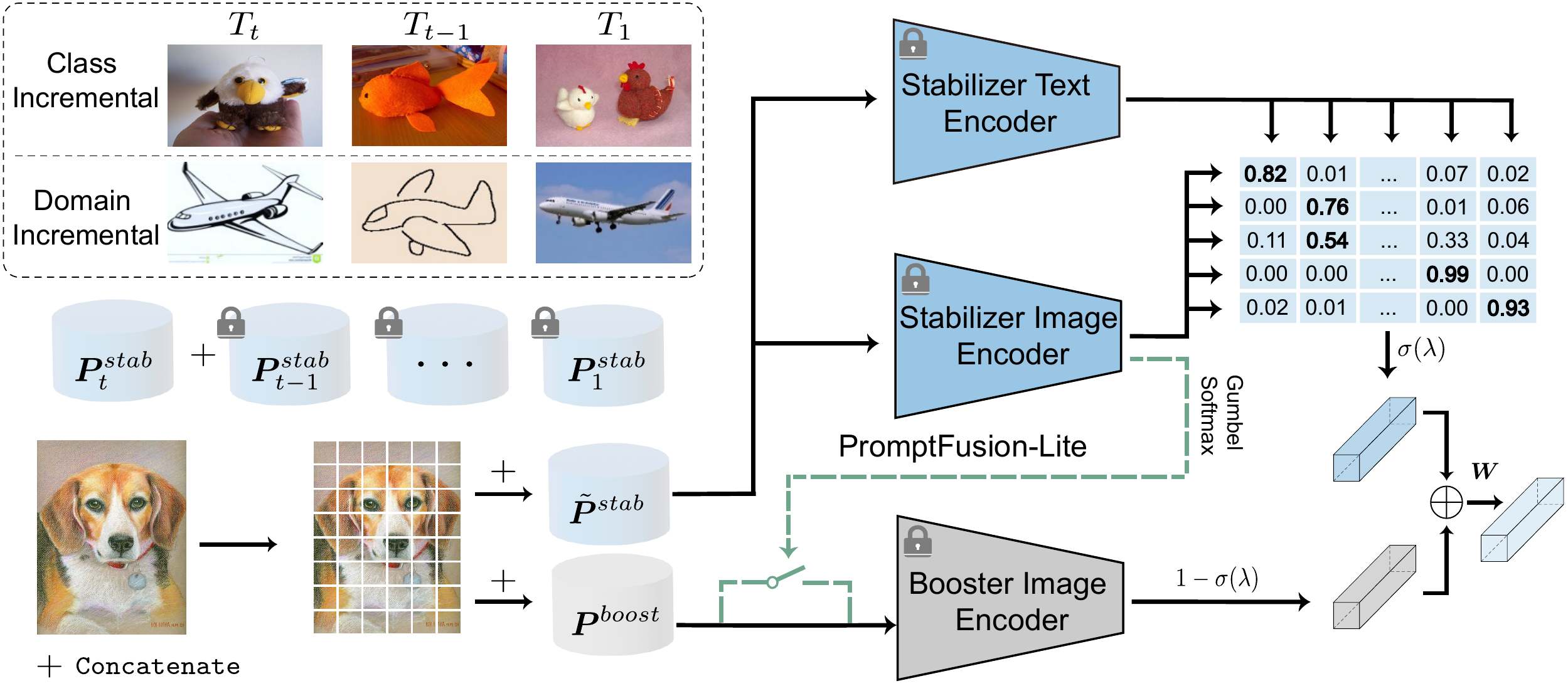}
   \caption{An overview of PromptFusion and PromptFusion-Lite. PromptFusion consists of two architectures, the \stab, and the \boo; PromptFusion-Lite has an additional decision module that adaptively determines whether to use the \boo on a per-input basis. 
   Note that both our proposed methods work in class incremental and domain incremental settings.}
   \label{fig:model}
\end{figure}

\noindent\textbf{PromptFusion with memory} With results from the pilot study, we confirm that stability and plasticity can be decoupled using the proposed \stab module and the \boo module. Based on this observation, we first present PromptFusion. The overview of PromptFusion is given in Figure~\ref{fig:model}. Formally, denote the \stab model as $S$ and the \boo model as 
$B$. Given an input image $\bm{x_i}$ with label $y_i$, $\bm{x_i}$ is first passed to the two modules $S$ and $B$. Their respective output $S(\bm{x_i})$ and $B(\bm{x_i})$ are then fused by the trainable parameter $\lambda$ through a weighted average. Due to using a memory buffer, the result is further element-wise multiplied by a weight mask $\bm{W}$ to balance old and new classes. Specifically, we would like old classes to be rectified and new classes to be weakened. Therefore, we divide the learning of $\bm{W}$ into two matrices, $\alpha$ and $\beta$ such that:
\begin{equation}
    \bm{W} = \text{Concat}[\frac{\theta}{\sigma(\beta)},\sigma(\alpha)], \theta = \frac{T_i}{2},
\end{equation}
 where $\sigma(\cdot) = \frac{1}{1+e^{-x}}$ is the sigmoid function and $\theta$ is a weight parameter scaled by the current task id $T_i$. Putting them together, the final output $\bm{z}_i$ is derived as:
\begin{equation}
    \bm{z}_i = \bm{W} \odot [(1 - \sigma(\lambda))S(\bm{x}_i) +\sigma(\lambda) B(\bm{x_i})],
\end{equation}
where $\odot$ indicates the element-wise multiplication operation. Further, following the idea in \cite{bahng2022clipaugment} that suggests CLIP can be enhanced with image prompts, we apply a similar approach to augment CoOp by inserting prompts $\Tilde{\bm{P}}^{stab}$ in the image patches. Finally, with $\texttt{CE}$ referring to the cross-entropy loss, the learning objective is:
\begin{equation}
\label{eqn:promptfusion objective}
    \min_{\bm \Theta} \sum_{i=1} \texttt{CE}(\bm{z}_i, y_i), \bm \Theta := \{\alpha, \beta, \bm{P}^{stab}, \bm{\Tilde{P}}^{stab}, \bm{P}^{boost} \}.
\end{equation}

\noindent\textbf{PromptFusion without memory} To align with state-of-the-art prompt-based methods that achieve promising results under the memory-free setting, we refer to the rehearsal approach~\cite{zhang2023slca} that generates old task statistics when training on the current task. Specifically, when learning each new task, we record the mean $\mu_k$ and covariance $\Sigma_k$ of the feature for each class $k$ and model the class feature as a Gaussian $\mathcal{N}(\mu_k, \Sigma_k)$. Then, when passing on to the next task, we first sample features from the distribution, and further finetune PromptFusion on the generated set. Since the set is composed of both features from the current and past tasks, it is relatively balanced. Therefore, the weight mask $\bm{W}$ is also accommodated accordingly:

\begin{equation}
    \bm{W} = \text{Concat}[\frac{1}{\sigma(\beta)},\sigma(\alpha)]
\end{equation}

\noindent\textbf{PromptFusion-Lite} While we can use both modules simultaneously and naively weighted sum their output, doing so would inevitably increase the computational cost. Based on existing works that demonstrate the efficacy of the \stab in continual learning, we seek an alternative approach, where we empirically run the \stab by default, and the use of the \boo is determined on-the-fly conditioned on inputs. It is worth pointing out that deriving on/off binary decisions is non-trivial, as it is non-differentiable. To mitigate this issue, we use a straight-through Gumbel-Softmax estimator~\cite{gumbelsoftmax}. Specifically, for each input image, we compute a decision entry as:
\begin{equation}
\begin{aligned}
\label{eqn:gumbel}
    M_{i,k}(v_i) &= \frac{\texttt{exp}{(\texttt{log}( \bm{F} (v_{i})_k + G_{i,k}) / \tau)}}{\sum_{j=1}^K \texttt{exp}{(\texttt{log}(\bm{F} (v_{i})_j + G_{i,j}) / \tau)}}, \\
\end{aligned}
\end{equation}
where $K=2$ is the number of categories, $G_i = - \log (-\log(U_i))$ is the Gumbel distribution in which $U_i \sim U(0,1)$, $v_i$ is the feature vector of the input image obtained from the image encoder of CLIP, $\bm{F}(\cdot)$ is an MLP consisting of two linear layers with the $\texttt{tanh}$ activation function that transforms the feature vector to a 2-dimensional output, and $\tau$ is a temperature parameter. Without loss of generality, we set $M_{i,0} = 1$ to indicate using the \boo and $M_{i,1}=1$ to indicate the opposite, such that the input $x_i$ is masked accordingly when passing to $B$. In contrast to  Eqn~\ref{eqn:promptfusion objective}, the learning objective for PromptFusion-Lite is now:

\begin{equation}
\label{eqn:promptfusionlite objective}
\begin{aligned}
    \min_{\bm \Theta} \sum_{i=1} \texttt{CE}&(\bm{z}_i, y_i) + \zeta ( \sum_{i=1} M_{i, 0} - \gamma)^2 + \delta \texttt{KD}(\bm{F}, \bm{F}'),\\
    \bm \Theta &:= \{\alpha, \beta, \bm{P}^{stab}, \bm{\Tilde{P}}^{stab}, \bm{P}^{boost}, \bm{F}\},
\end{aligned}
\end{equation} 
where the second term refers to a usage penalty that limits the activation of the \boo and the last term refers to the knowledge distillation of $\bm{F}$ before and after training on the new task, intending to mitigate its forgetting. Here $\bm{F}'$ refers to a frozen copy of $\bm{F}$ after finishing training on the current task. $\gamma$ is a predefined target for the fraction of times the \boo is expected to use and $\zeta$ and $\delta$ are weight parameters controlling the contribution to the total loss. The full picture of the algorithm is given in the appendix.

\section{Experiments}
\subsection{Experimental Setup}
\noindent\textbf{Datasets} Experiments are conducted on four popular benchmark datasets of continual learning, namely Cifar100, CUB200, Imagenet-R, and Core50, where Cifar100, CUB200, and Imagenet-R are evaluated under the class-incremental setting, and Core50 is evaluated under the domain-incremental setting. Cifar100 is a relatively simple dataset containing 100 classes. CUB200 is composed of 200 subcategories belonging to birds. Imagenet-R is also composed of 200 classes that consist of data from different styles such as cartoons, graffiti, and origami. It is considered one of the most difficult datasets for class-incremental learning as semantic and covariate shifts occur. Throughout the experiments, all the above datasets are split into 10 tasks with an even number of classes in each task. Core50, on the other hand, is a popular dataset for domain-incremental learning that consists of 50 objects from 11 domains. In particular, 8 of them are used for training and the rest 3 for testing. None of the images in the 3 domains are seen during training. For all three datasets, we use class orders the same as \cite{learningtoprompt, wang2022dualprompt} for a fair comparison.

\noindent\textbf{Evaluation Metrics} Following conventional settings, we report the Average Accuracy after training on all tasks to evaluate the overall continual learning ability of the tested method. Formally, let $R_{T, i}$ be the classification accuracy of task $T_i$ after training on task $T_T$, then the Average Accuracy $A_T$ is defined as
\begin{equation}
\label{eqn:average accuracy}
A_T = \frac{1}{T}\sum_{i=1}^{T} R_{T, i}.
\end{equation}

\noindent\textbf{Implementation Details} We adopt a ViT-B-16 backbone for both CoOp and VPT and leave it frozen during the training phase. For the prompt size, $M$ in CoOp and $p$ in VPT are set to 30. The prompt length for augmentating CoOp is set to 40. All prompts are trained using the mini-batch AdamW optimizer with a learning rate of 0.002 on Split-Cifar100 and Split-CUB200, and a learning rate of 0.003 on Split-Imagenet-R and Core50, equipped with a cosine annealing scheduler. Considering their diverse difficulties, Split-Cifar100 and Split-CUB200 are trained for 3 epochs for every task while Split-ImageNet-R and Core50 are trained for 5 epochs. 
% $\tau$ in Eqn~\ref{eqn:gumbel} is set to 3, and $\zeta$ , $\gamma$ and $\delta$ in Eqn~\ref{eqn:promptfusionlite objective} are set to 0.2, 0.4 and 0.1 respectively.

\begin{table*}[t]
\begin{center}
\caption{Results on the Split-Cifar100, Split-CUB200, Split-Imagenet-R and Core50 datasets. Split-Cifar100, Split-Imagenet-R and Split-CUB200 are class-incremental learning datasets, and Core50 is a domain-incremental learning dataset. Compared methods are grouped based on memory size. * denotes that the results are obtained through our PyTorch re-implementation~\cite{sun2023pilot}.
} 
\resizebox{\linewidth}{!}{
\begin{tabular}{lcccccccc}
\toprule 
 \multirow{2}{*}{\textbf{Method}} & \multicolumn{2}{c}{\textbf{Split-Cifar100}}
 &  \multicolumn{2}{c}{\textbf{Split-CUB200}} &  \multicolumn{2}{c}{\textbf{Split-Imagenet-R}} & \multicolumn{2}{c}{\textbf{Core50}} \\ 
&Buffer Size &  Avg Acc   & Buffer Size & Avg Acc & Buffer Size  &  Avg Acc & Buffer Size & Avg Acc \\
\cmidrule(lr){1-1} \cmidrule(lr){2-3}  \cmidrule(lr){4-5}  \cmidrule(lr){6-7} \cmidrule(lr){8-9}
 BiC \cite{wu2019bic}  &\multirow{9}{*}{5000} & 81.4  &\multirow{9}{*}{1000} & 81.9 &\multirow{9}{*}{5000}&  64.6 &\multirow{9}{*}{5000} & 79.3\\
 GDumb \cite{prabhu2020gdumb} & & 81.7&& 61.8 &  & 65.9& & 74.9\\
 DER++ \cite{yan2021dynamically} & & 83.9 & & 77.4&  & 66.7 &  & 79.7\\
 C$\text{o}^2$L \cite{cha2021co2l} &  &82.5 &  &\textcolor{gray}{\emph{N.A.}}  & &65.9 & &79.8\\
 L2$\text{P}^{*}$ \cite{learningtoprompt} & &85.9 &&82.3&  &70.8 &  &85.1\\
 S-Prompt* \cite{wang2022sprompt} & &  \textcolor{gray}{\emph{N.A.}}&  &\textcolor{gray}{\emph{N.A.}} & &\textcolor{gray}{\emph{N.A.}} & &92.1\\
 DualPromp$\text{t}^{*}$ \cite{wang2022dualprompt}  & &87.5 &  & 83.9 &&65.2&&87.2\\
  PromptFusion &  & \textbf{88.5}  & & \textbf{85.1}&  &\textbf{82.2} &  &\textbf{93.7}\\
PromptFusion-Lite &  & 87.3   & & 82.7  & & 81.5 & & 93.2\\
\bottomrule
\end{tabular}}
\label{table:main_results}
\end{center}
\end{table*}

\subsection{Performance Comparison to State-of-the-art}
We compare against both classical and most recent state-of-the-art continual learning methods: EWC~\cite{ewc}, LwF~\cite{lwf}, BiC~\cite{wu2019bic}, GDumb~\cite{prabhu2020gdumb}, DER++~\cite{yan2021dynamically}, C$\text{o}^2$L~\cite{cha2021co2l}, L2P~\cite{learningtoprompt}, S-Prompt~\cite{wang2022sprompt}, DualPrompt~\cite{wang2022dualprompt} and CODA-Prompt~\cite{smithcodaprompt}. In particular, DER++ is the best-performing non-prompt-based method, while L2P, S-Prompt, DualPrompt, and CODA-Prompt are all prompt-based methods. Since S-Prompt is specifically designed for domain-incremental learning, we only report its performance on the Core50 dataset.

The results on Split-Cifar100, Split-CUB200, Split-Imagenet-R and and Core50 are shown in Table \ref{table:main_results}, where PromptFusion outperforms all other alternatives. 
\setlength{\intextsep}{5pt}
\begin{wrapfigure}[15]{r}{0.5\textwidth}
  \centering
   \includegraphics[width=0.5\columnwidth,height=0.4\columnwidth]{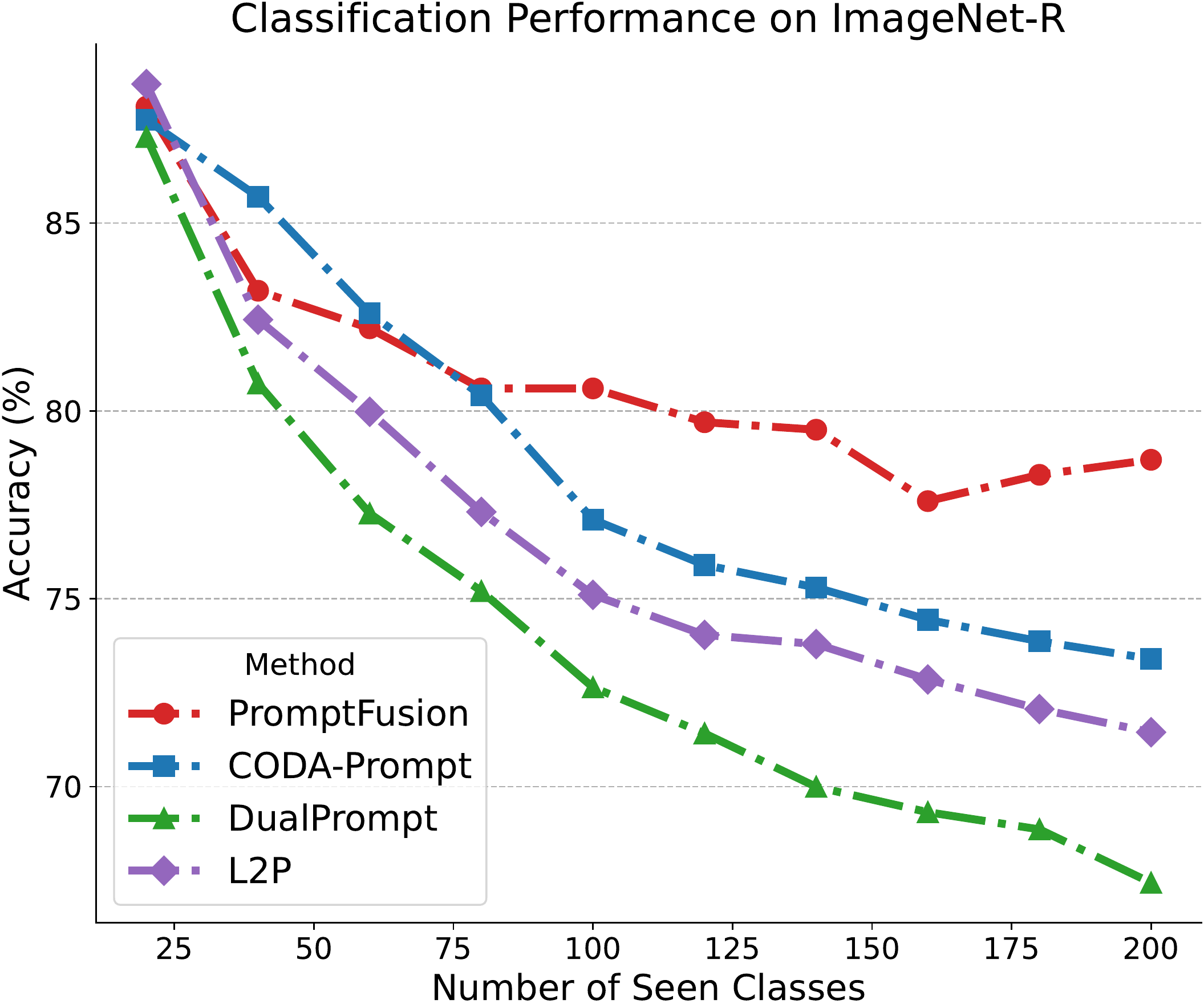}
   \caption{Comparison on Split-Imagenet-R with no memory.}
   \label{fig:imagenetr_0memory}
\end{wrapfigure}
For Split-Cifar100, PromptFusion reaches an Avg Acc of 88.5\%. The performance is 2.6\% and 1\% higher than L2P and DualPrompt respectively. For Split-CUB200, PromptFusion outperforms L2P and DualPrompt by 2.8\% and 1.2\%. Considering that Cifar100 and CUB200 are relatively simple, the improvement is significant. For Split-Imagenet-R, PromptFusion can achieve an Avg Acc of 82.2\%, significantly surpassing other methods. Similarly, PromptFusion-Lite also achieves competitive performances with less computational overhead than PromptFusion, as will be discussed in the next subsection.

Here, the above results are obtained using a memory buffer. Since the current paradigm for prompt-based methods also focuses on the memory-free situation, we test PromptFusion on Split-Imagenet-R with 0 memory. As is shown in Fig~\ref{fig:imagenetr_0memory}, PromptFusion outperforms all other methods. In particular, when compared against CODA-Prompt, the best prompt-based approach, PromptFusion outperforms it by 5.3\%. A major reason for this performance gain is that incorporating the \stab makes our method robust against intra-class variations, as discussed in Section~\ref{sec:pilot study}. 

\setlength{\intextsep}{0pt}
\begin{wraptable}{r}{0.4\columnwidth}
\centering
\caption{GFLOPs of prompt-based methods on Split-Imagnet-R.} 
\resizebox{0.4\columnwidth}{!}{
\begin{tabular}{lcc}
\toprule 
 \multirow{2}{*}{\textbf{Method}} & \multicolumn{2}{c}{\textbf{Split-Imagnet-R}}
  \\ 
& Avg Acc & GFLOPs     \\
\cmidrule(lr){1-1} \cmidrule(lr){2-3}   
 L2P \cite{prabhu2020gdumb} & 70.8  & 38.9 \\
 DualPrompt \cite{yan2021dynamically} & 65.2 & 35.2 \\
PromptFusion & \textbf{82.2} &  42.6 \\
PromptFusion-Lite & 81.5 &  34.4 \\
\bottomrule
\end{tabular}}
\label{table:computational comparison}
\end{wraptable}

While Split-Cifar100, Split-CUB200, and Split-Imagenet-R are all commonly used datasets in the class-incremental learning scenario, we also test PromptFusion and PromptFusion-Lite under the domain-incremental learning scenario on the Core50 dataset. Notably, under such setting, both L2P and DualPrompt perform inferior to S-Prompt, as their main focus is on class-incremental learning. Nevertheless, our methods still outperform S-Prompt by 1.1\% and 1.3\% respectively, indicating that the capability of them is not restricted to any specific type of continual learning setting. Considering that S-Prompt is also based on CoOp, this is strong evidence that the success of our approach is not simply from this specific module.

\subsection{GFLOPs Comparisons to State-of-the-art}

\noindent{While} computational concerns might be raised about our approaches, we show that they are similar to other prompt-based methods. This is because for L2P and DualPrompt, an additional raw pre-trained ViT is used to extract the [CLS] feature of the input to select the optimal prompts from the prompt pool. Consequently, as is shown in Table~\ref{table:computational comparison}, PromptFusion only requires about $15\%$ more GFLOPs than L2P and DualPrompt. Moreover, by incorporating the adaptive selection module, PromptFusion-Lite's GFLOPs requirement can even fall below that of L2P and DualPrompt. In particular, on Split-Imagnet-R, PromptFusion-Lite achieves an Avg Acc of 0.7\% less than PromtFusion with 19.2\% fewer GFLOPs. 

\subsection{Effect of memory size}

\setlength{\intextsep}{5pt}
\begin{wrapfigure}[12]{r}{0.55\textwidth}
  % \centering
  \centering
   \includegraphics[width=0.55\textwidth,height=0.28\textwidth]{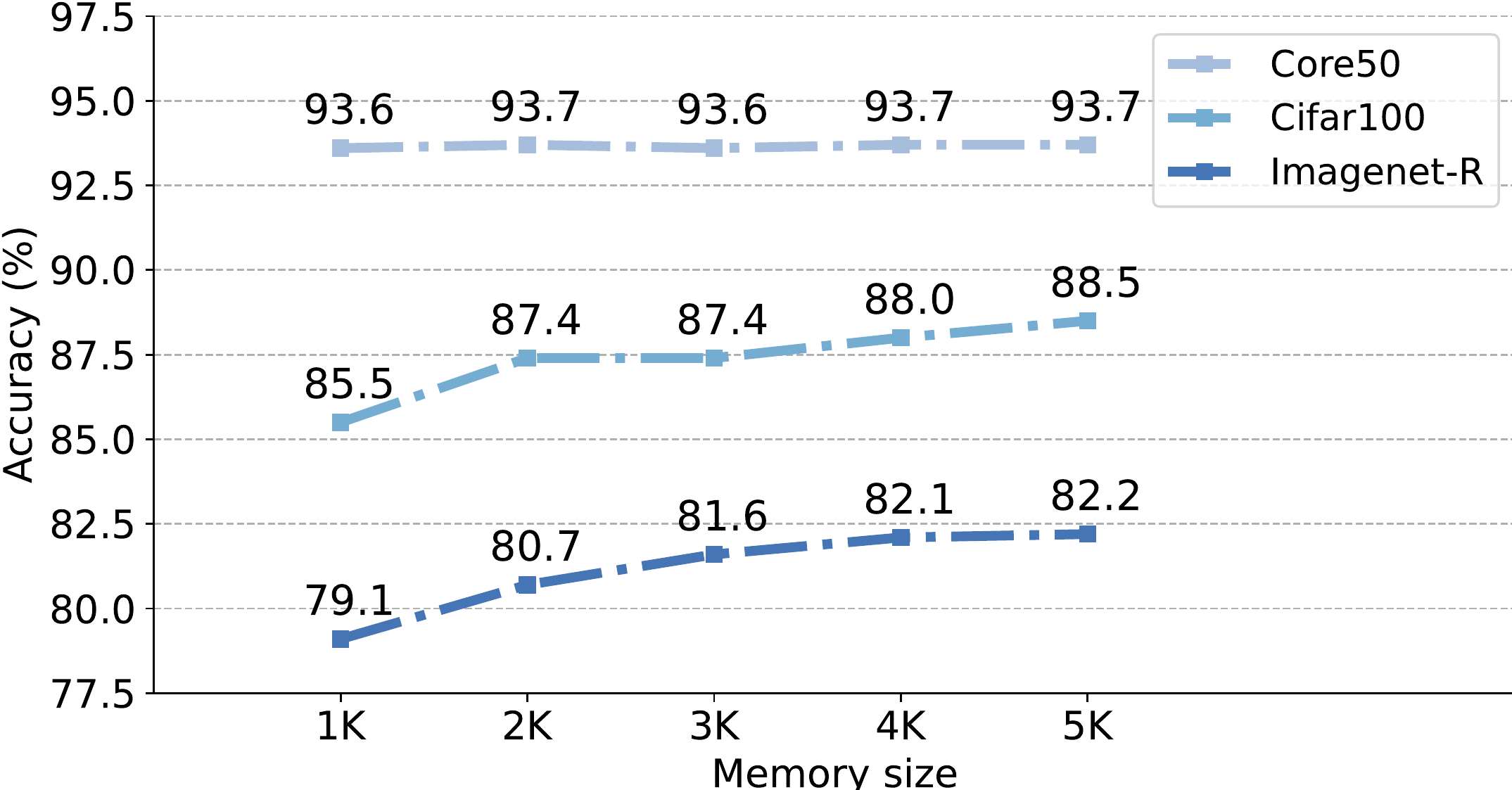}
   \caption{Average Acc for different memory sizes on Split-Cifar100, Split-Imagenet-R, and Core50.}
   \label{fig:memory}
\end{wrapfigure}
While PromptFusion shows competitive performance in the memory-free setting, we would also like to test how it behaves for different buffer sizes. We analyze sizes of 1K, 2K, 3K, 4K, and 5K and report the results in Figure~\ref{fig:memory}. In general, the effect of buffer size on the overall performance is rather small. Specifically on Core50, the buffer size does not significantly affect the performance of PromptFusion. This is unsurprising as the test data for Core50 is a constant set that has not been seen during training. Therefore the only factor influencing the performance is the model's capability of accumulating domain-invariant knowledge from sequential training on the training set, which PromptFusion demonstrates to excel at.

\subsection{Ablation Study}
\noindent\textbf{Weight $\lambda$, Mask $\bm{W}$ and CoOp Augmentation} We report in Table~\ref{tab:weight ablation} the ablation study on weight $\lambda$, mask $\bm{W}$ and CoOp augmentation, and the results 
\begin{wraptable}{r}{0.5\columnwidth}
\tabcolsep = 0.12cm
\centering
\caption{Ablation on different components of PromptFusion.}
\resizebox{0.9\linewidth}{!}{
\begin{tabular}{cccc }
\toprule
Mask $\bm{W}$ &  $\lambda$& Augmentation  &  
\textbf{Split-Cifar100}  \\
  \cmidrule(lr){1-1}  \cmidrule(lr){2-2} \cmidrule(lr){3-3}  \cmidrule(lr){4-4} 

\ding{53}&  \ding{51} & \ding{51} &84.2  \\
\ding{51}&  \ding{53} &\ding{51}&87.4   \\
\ding{51}&  \ding{51} & \ding{53}   & 87.1   \\
\ding{51}&  \ding{51} & \ding{51}   & \textbf{88.5}   \\
\bottomrule
\end{tabular}}
\label{tab:weight ablation}
\end{wraptable}
show that all pieces are significant to the success of our methods. Specifically, for Mask $\bm{W}$, its effect on Split-Cifar100 is much higher than that on Split-Imagenet-R. We posit that this is because new and old classes in the Imagenet-R dataset are more diverse as semantic and covariate shifts occur, resulting in weaker inter-class interference. As for $\lambda$, excluding it in PromptFusion means a simple summation of the two outcomes, and results in Table~\ref{tab:weight ablation} demonstrate it to be sub-optimal. As discussed in Section~\ref{sec:pilot study}, we expect different values of $\lambda$ depending on the dataset being assessed. Indeed, experiments show that $\lambda=1.06$ for Split-Cifar100 and $\lambda =0.01$ for Split-Imagenet-R. Finally, as introduced in Section~\ref{sec:prompt fusion}, CoOp is augmented by incorporating another set of image prompts in addition to the language prompt, and the performance would increase by 1.4\%, with only 0.02M extra trainable parameters.

\noindent\textbf{Usage Penalty and Knowledge Distillation} We further conduct an ablation 
\setlength{\intextsep}{0pt}
\begin{wraptable}{r}{0.5\columnwidth}
\centering
\caption{Ablation on PromptFusion-Lite objective.}
\resizebox{0.9\linewidth}{!}{
\begin{tabular}{cccc}
\toprule
\multirow{2}{*}{Usage Penalty} & \multirow{2}{*}{Knowl. Distill.} & \multicolumn{2}{c}{\textbf{Split-Cifar100}} \\ 
 & & Avg Acc & GFLOPs  \\
  \cmidrule(lr){1-1}  \cmidrule(lr){2-2} \cmidrule(lr){3-4}   
\ding{53}&  \ding{51}  &\textbf{88.2} &  42.6\\
\ding{51}&  \ding{53} & 87.0 & 38.8 \\
\ding{51}&  \ding{51}   & 87.3 &  \textbf{38.3}\\
\bottomrule
\end{tabular}}
\label{tab:promptfusion-lite objective ablation}
\end{wraptable}
study on PromptFusion-Lite targeting its objective function. Specifically, we would like to analyze the effect of the second and third terms in Eqn~\ref{eqn:promptfusionlite objective}. As is shown in Tab~\ref{tab:promptfusion-lite objective ablation}, without using the usage penalty term, PromptFusion-Lite would almost always choose to activate the \boo, as evidenced by it having the same GFLOPs as PromptFusion. Without using knowledge distillation on 
$\mathbf{F}$, on the other hand, will result in a 0.3\% performance degradation and slightly higher GFLOPs. Therefore, we conclude that both pieces are significant to the overall success of PromptFusion-Lite.

\noindent\textbf{Prompt Length} We also examine how prompt length affects the overall performance and the results are reported in Table~\ref{tab:prompt length}. As is shown, our choice of $M=30$ and $p=30$ produces the best Average Acc. This would require a total of 1.66M trainable parameters on Split-Cifar100, which is infinitesimal compared to other approaches.

\begin{table}[h]
\centering
\caption{Ablation on prompt length $M$ and $p$ for PromptFusion}
\resizebox{0.7\linewidth}{!}{
 \begin{tabular}{c  c || c c}
\toprule
 Text Prompt & \textbf{Split-Cifar100} & Image Prompt & \textbf{Split-Cifar100} \\
 \midrule
$M=20$ &   88.1   &  $p=20$   &  87.9 \\
$M=40$  &   88.2   & $p=40$     &  88.1 \\
\midrule
$M=30$ &   \textbf{88.5}  & $p=30$     &  \textbf{88.5}  \\
\bottomrule
\end{tabular}}

\label{tab:prompt length}

\end{table}

\section{Conclusion}
In this paper, we introduced a dual architecture design PromptFusion for tackling the stability-plasticity dilemma in continual learning. PromptFusion is built on top of a \stab module instantiated with CoOp and a \boo module instantiated with VPT, that decouples stability and plasticity into two independent problems. Furthermore, to reduce the computational overhead caused by the additional architecture, we proposed PromptFusion-Lite that leverages the \stab as a base model, and adaptively decides on-the-fly whether to activate the \boo conditioned on the input sample. Extensive experiments showed that our approach achieved state-of-the-art results in class-incremental and domain-incremental learning under both memory and memory-free settings with PromptFusion-Lite successfully reducing more than 10\% computational cost. 

\section*{Acknowledgement} 
This project was supported by NSFC under Grant No. 62102092.

\bibliographystyle{splncs04}
\bibliography{main}

\begin{thebibliography}{10}
\providecommand{\url}[1]{\texttt{#1}}
\providecommand{\urlprefix}{URL }
\providecommand{\doi}[1]{https://doi.org/#1}

\bibitem{abraham2005memory}
Abraham, W.C., Robins, A.: Memory retention--the synaptic stability versus plasticity dilemma. Trends in neurosciences  (2005)

\bibitem{mas}
Aljundi, R., Babiloni, F., Elhoseiny, M., Rohrbach, M., Tuytelaars, T.: Memory aware synapses: Learning what (not) to forget. In: ECCV (2018)

\bibitem{bahng2022clipaugment}
Bahng, H., Jahanian, A., Sankaranarayanan, S., Isola, P.: Exploring visual prompts for adapting large-scale models. arXiv preprint arXiv:2203.17274  (2022)

\bibitem{il2m}
Belouadah, E., Popescu, A.: Il2m: Class incremental learning with dual memory. In: ICCV (2019)

\bibitem{cha2021co2l}
Cha, H., Lee, J., Shin, J.: Co2l: Contrastive continual learning. In: ICCV (2021)

\bibitem{mpa}
Chen, H., Wu, Z., Jiang, Y.G.: Multi-prompt alignment for multi-source unsupervised domain adaptation. NeurIPS  (2023)

\bibitem{continualsurvey1}
De~Lange, M., Aljundi, R., Masana, M., Parisot, S., Jia, X., Leonardis, A., Slabaugh, G., Tuytelaars, T.: A continual learning survey: Defying forgetting in classification tasks. TPAMI  (2021)

\bibitem{lwm}
Dhar, P., Singh, R.V., Peng, K.C., Wu, Z., Chellappa, R.: Learning without memorizing. In: CVPR (2019)

\bibitem{douillard2022dytox}
Douillard, A., Ram{\'e}, A., Couairon, G., Cord, M.: Dytox: Transformers for continual learning with dynamic token expansion. In: CVPR (2022)

\bibitem{fini2022self}
Fini, E., Da~Costa, V.G.T., Alameda-Pineda, X., Ricci, E., Alahari, K., Mairal, J.: Self-supervised models are continual learners. In: CVPR (2022)

\bibitem{dapl}
Ge, C., Huang, R., Xie, M., Lai, Z., Song, S., Li, S., Huang, G.: Domain adaptation via prompt learning. arXiv preprint arXiv:2202.06687  (2022)

\bibitem{continualsurvey2}
Hadsell, R., Rao, D., Rusu, A.A., Pascanu, R.: Embracing change: Continual learning in deep neural networks. Trends in cognitive sciences  (2020)

\bibitem{Resnet}
He, K., Zhang, X., Ren, S., Sun, J.: Deep residual learning for image recognition. In: CVPR (2016)

\bibitem{hou2019lucir}
Hou, S., Pan, X., Loy, C.C., Wang, Z., Lin, D.: Learning a unified classifier incrementally via rebalancing. In: CVPR (2019)

\bibitem{cpg}
Hung, C.Y., Tu, C.H., Wu, C.E., Chen, C.H., Chan, Y.M., Chen, C.S.: Compacting, picking and growing for unforgetting continual learning. Advances in Neural Information Processing Systems  \textbf{32} (2019)

\bibitem{gumbelsoftmax}
Jang, E., Gu, S., Poole, B.: Categorical reparameterization with gumbel-softmax. arXiv preprint arXiv:1611.01144  (2016)

\bibitem{vpt}
Jia, M., Tang, L., Chen, B.C., Cardie, C., Belongie, S., Hariharan, B., Lim, S.N.: Visual prompt tuning. In: ECCV (2022)

\bibitem{ju2021videoprompting}
Ju, C., Han, T., Zheng, K., Zhang, Y., Xie, W.: Prompting visual-language models for efficient video understanding. arXiv preprint arXiv:2112.04478  (2021)

\bibitem{ewc}
Kirkpatrick, J., Pascanu, R., Rabinowitz, N., Veness, J., Desjardins, G., Rusu, A.A., Milan, K., Quan, J., Ramalho, T., Grabska-Barwinska, A., et~al.: Overcoming catastrophic forgetting in neural networks. PNAS  (2017)

\bibitem{alexnet}
Krizhevsky, A., Sutskever, I., Hinton, G.E.: Imagenet classification with deep convolutional neural networks. In: NeurIPS (2012)

\bibitem{kumaran2016learning}
Kumaran, D., Hassabis, D., McClelland, J.L.: What learning systems do intelligent agents need? complementary learning systems theory updated. Trends in cognitive sciences  (2016)

\bibitem{lester2021power}
Lester, B., Al-Rfou, R., Constant, N.: The power of scale for parameter-efficient prompt tuning. arXiv preprint arXiv:2104.08691  (2021)

\bibitem{li2021prefix}
Li, X.L., Liang, P.: Prefix-tuning: Optimizing continuous prompts for generation. arXiv preprint arXiv:2101.00190  (2021)

\bibitem{lwf}
Li, Z., Hoiem, D.: Learning without forgetting. TPAMI  (2017)

\bibitem{liu2021promptsurvey}
Liu, P., Yuan, W., Fu, J., Jiang, Z., Hayashi, H., Neubig, G.: Pre-train, prompt, and predict: A systematic survey of prompting methods in natural language processing. arXiv preprint arXiv:2107.13586  (2021)

\bibitem{lopez2017gradient}
Lopez-Paz, D., Ranzato, M.: Gradient episodic memory for continual learning. NeurIPS  (2017)

\bibitem{lu2022promptdistribution}
Lu, Y., Liu, J., Zhang, Y., Liu, Y., Tian, X.: Prompt distribution learning. In: CVPR (2022)

\bibitem{bayesianpromptlearning}
Mahdi~Derakhshani, M., Sanchez, E., Bulat, A., Guilherme Turrisi~da Costa, V., Snoek, C.G., Tzimiropoulos, G., Martinez, B.: Bayesian prompt learning for image-language model generalization. ICCV  (2023)

\bibitem{continualsurvey3}
Mai, Z., Li, R., Jeong, J., Quispe, D., Kim, H., Sanner, S.: Online continual learning in image classification: An empirical survey. Neurocomputing  (2022)

\bibitem{mallya2018packnet}
Mallya, A., Lazebnik, S.: Packnet: Adding multiple tasks to a single network by iterative pruning. In: CVPR (2018)

\bibitem{cls}
McClelland, J.L., McNaughton, B.L., O'Reilly, R.C.: Why there are complementary learning systems in the hippocampus and neocortex: insights from the successes and failures of connectionist models of learning and memory. Psychological review  (1995)

\bibitem{mermillod2013stability}
Mermillod, M., Bugaiska, A., Bonin, P.: The stability-plasticity dilemma: Investigating the continuum from catastrophic forgetting to age-limited learning effects (2013)

\bibitem{mirzadeh2022architecture}
Mirzadeh, S.I., Chaudhry, A., Yin, D., Nguyen, T., Pascanu, R., Gorur, D., Farajtabar, M.: Architecture matters in continual learning. arXiv preprint arXiv:2202.00275  (2022)

\bibitem{prabhu2020gdumb}
Prabhu, A., Torr, P.H., Dokania, P.K.: Gdumb: A simple approach that questions our progress in continual learning. In: ECCV (2020)

\bibitem{radford2021clip}
Radford, A., Kim, J.W., Hallacy, C., Ramesh, A., Goh, G., Agarwal, S., Sastry, G., Askell, A., Mishkin, P., Clark, J., et~al.: Learning transferable visual models from natural language supervision. In: ICML (2021)

\bibitem{icarl}
Rebuffi, S.A., Kolesnikov, A., Sperl, G., Lampert, C.H.: icarl: Incremental classifier and representation learning. In: CVPR (2017)

\bibitem{fasterrcnn}
Ren, S., He, K., Girshick, R., Sun, J.: Faster r-cnn: Towards real-time object detection with region proposal networks. In: NeurIPS (2015)

\bibitem{progressive}
Rusu, A.A., Rabinowitz, N.C., Desjardins, G., Soyer, H., Kirkpatrick, J., Kavukcuoglu, K., Pascanu, R., Hadsell, R.: Progressive neural networks. arXiv preprint arXiv:1606.04671  (2016)

\bibitem{shin2017generativereplay}
Shin, H., Lee, J.K., Kim, J., Kim, J.: Continual learning with deep generative replay. NeurIPS  (2017)

\bibitem{smithcodaprompt}
Smith, J.S., Karlinsky, L., Gutta, V., Cascante-Bonilla, P., Kim, D., Arbelle, A., Panda, R., Feris, R., Kira, Z.: Coda-prompt: Continual decomposed attention-based prompting for rehearsal-free continual learning. In: CVPR (2023)

\bibitem{sun2023pilot}
Sun, H.L., Zhou, D.W., Ye, H.J., Zhan, D.C.: Pilot: A pre-trained model-based continual learning toolbox. arXiv preprint arXiv:2309.07117  (2023)

\bibitem{van2019three}
Van~de Ven, G.M., Tolias, A.S.: Three scenarios for continual learning. arXiv preprint arXiv:1904.07734  (2019)

\bibitem{von2019hyper}
Von~Oswald, J., Henning, C., Sacramento, J., Grewe, B.F.: Continual learning with hypernetworks. arXiv preprint arXiv:1906.00695  (2019)

\bibitem{wang2022sprompt}
Wang, Y., Huang, Z., Hong, X.: S-prompts learning with pre-trained transformers: An occam's razor for domain incremental learning. arXiv preprint arXiv:2207.12819  (2022)

\bibitem{wang2022dualprompt}
Wang, Z., Zhang, Z., Ebrahimi, S., Sun, R., Zhang, H., Lee, C.Y., Ren, X., Su, G., Perot, V., Dy, J., et~al.: Dualprompt: Complementary prompting for rehearsal-free continual learning. arXiv preprint arXiv:2204.04799  (2022)

\bibitem{learningtoprompt}
Wang, Z., Zhang, Z., Lee, C.Y., Zhang, H., Sun, R., Ren, X., Su, G., Perot, V., Dy, J., Pfister, T.: Learning to prompt for continual learning. In: CVPR (2022)

\bibitem{wei2021pretrained}
Wei, C., Xie, S.M., Ma, T.: Why do pretrained language models help in downstream tasks? an analysis of head and prompt tuning. In: NeurIPS (2021)

\bibitem{wu2021striking}
Wu, G., Gong, S., Li, P.: Striking a balance between stability and plasticity for class-incremental learning. In: ICCV. pp. 1124--1133 (2021)

\bibitem{wu2019bic}
Wu, Y., Chen, Y., Wang, L., Ye, Y., Liu, Z., Guo, Y., Fu, Y.: Large scale incremental learning. In: CVPR (2019)

\bibitem{wu2024building}
Wu, Z., Weng, Z., Peng, W., Yang, X., Li, A., Davis, L.S., Jiang, Y.G.: Building an open-vocabulary video clip model with better architectures, optimization and data. TPAMI  (2024)

\bibitem{yan2021dynamically}
Yan, S., Xie, J., He, X.: Der: Dynamically expandable representation for class incremental learning. In: CVPR (2021)

\bibitem{si}
Zenke, F., Poole, B., Ganguli, S.: Continual learning through synaptic intelligence. In: ICML (2017)

\bibitem{zhang2023slca}
Zhang, G., Wang, L., Kang, G., Chen, L., Wei, Y.: Slca: Slow learner with classifier alignment for continual learning on a pre-trained model. In: ICCV (2023)

\bibitem{coop}
Zhou, K., Yang, J., Loy, C.C., Liu, Z.: Learning to prompt for vision-language models. IJCV  (2022)

\end{thebibliography}
\end{document}